\documentclass{article}
\usepackage{spconf,amsmath,graphicx,microtype}
\usepackage{array,multirow}

\newcommand\blfootnote[1]{%
  \begingroup
  \renewcommand\thefootnote{}\footnote{#1}%
  \addtocounter{footnote}{-1}%
  \endgroup
}
\title{TOWARDS END-TO-END INTEGRATION OF DIALOG HISTORY FOR IMPROVED SPOKEN LANGUAGE UNDERSTANDING}
\name{\begin{tabular}{c} Vishal Sunder$^{* 1}$, Samuel Thomas$^{2}$, Hong-Kwang J. Kuo$^{2}$, Jatin Ganhotra$^{2}$, \\ Brian Kingsbury$^{2}$, Eric Fosler-Lussier$^{1}$ \end{tabular}}
\address{\begin{tabular}{c} $^{1}$ The Ohio State University, Columbus, OH, USA \\ $^{2}$ IBM Research AI, Yorktown Heights, NY, USA \end{tabular}}
\begin{document}
%
\maketitle
\begin{abstract}
Dialog history plays an important role in spoken language understanding (SLU) performance in a dialog system. For end-to-end (E2E) SLU, previous work has used dialog history in text form, which makes the model dependent on a cascaded automatic speech recognizer (ASR). This rescinds the benefits of an E2E system which is intended to be compact and robust to ASR errors. In this paper, we propose a hierarchical conversation model that is capable of directly using dialog history in speech form, making it fully E2E. We also distill semantic knowledge from the available gold conversation transcripts by jointly training a similar text-based conversation model with an explicit tying of acoustic and semantic embeddings. We also propose a novel technique that we call DropFrame to deal with the long training time incurred by adding dialog history in an E2E manner. On the HarperValleyBank dialog dataset, our E2E history integration outperforms a history independent baseline by 7.7\% absolute F1 score on the task of dialog action recognition. Our model performs competitively with the state-of-the-art history based cascaded baseline, but uses 48\% fewer parameters. In the absence of gold transcripts to fine-tune an ASR model, our model outperforms this baseline by a significant margin of 10\% absolute F1 score.
\end{abstract}
\section{Introduction}
\label{sec:intro}
\blfootnote{*Work done during an internship at IBM}

Traditionally, spoken dialog systems (SDS) comprise an automatic speech recognition (ASR) component followed by a natural language understanding (NLU) component \cite{tur2002improving, hakkani2006beyond, henderson2012discriminative, huang2019adapting}. In recent times, end-to-end (E2E) spoken language understanding (SLU) has gained popularity for SDS \cite{rongali2020exploring, chung2021splat, morais2021end, denisov2020pretrained}. The reason is that state-of-the-art models for ASR and NLU can be inherently large and cascading them leads to an even larger overall model size, which may make them difficult to deploy, especially for on-device applications. Also, ASR errors can degrade the performance of a cascaded SLU system. In contrast, E2E SLU models are compact and more robust to ASR errors as they process speech directly.


It is also well known that a dialog system's NLU performance benefits from using the entire dialog history instead of just the utterance to be labeled. Dialog history helps particularly by resolving ambiguities and co-references. This idea has been explored extensively in written/typed dialog systems \cite{bothe2018context,raheja2019dialogue,colombo2020guiding}. Some recent work has explored ways of incorporating dialog history for intent identification in E2E SLU. However, a limitation here is that the features for the dialog history are extracted from a text encoder which makes the model dependent on a cascaded ASR.

\begin{figure}
    \hfill
    \centering
    \centerline{\includegraphics[width=0.9\columnwidth]{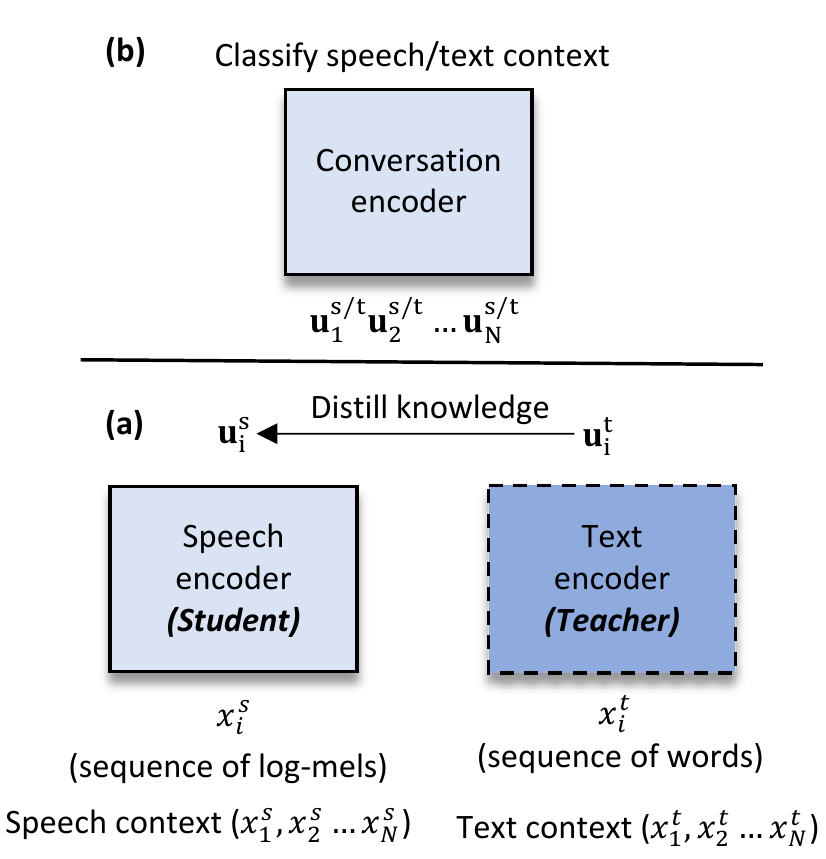}}
\caption{Model overview. \textbf{(a)} Modality specific utterance encoders; \textbf{(b)} Modality agnostic conversation encoder. The text encoder acts as a teacher for the speech encoder during training and is discarded at test time.}
\label{fig:model_over}
\end{figure}

In particular, Tomashenko et al. \cite{tomashenko2020dialogue} use text from the previous system prompt as history. In an extension to this, Ganhotra et al. \cite{ganhotra2021integrating} use the entire conversation as history which is a transcript decoded from an ASR, making it prone to ASR errors and also not fully E2E. ASR performance has also been shown to improve by integrating dialog history \cite{kim2019gated} and intent representations \cite{ray2021listen}. 

In this paper, we propose a conversation model with a full E2E integration of dialog history that is significantly more compact than its cascaded counterpart. Our conversation model is hierarchical, where we have an utterance level speech encoder at the lower level and a conversation encoder at the top level. 


To further improve the performance, we use a teacher-student training framework to distill knowledge from the semantically richer BERT encoder using available transcriptions. We jointly train a BERT based conversation model along with the speech based model with weight sharing at the conversation level between modalities (see Figure \ref{fig:model_over}). Further, speech embeddings are tied with BERT embeddings via Euclidean loss \cite{huang2020leveraging} and a novel use of Contrastive loss \cite{zhang2021cross}.


Using dialog history in an E2E manner can lead to an increased training time. We propose a technique called DropFrame to deal with this issue.
Similar to dropout \cite{srivastava2013improving}, this technique drops out random speech frames during training. In addition to a reduced training time, this technique also helps to improve SLU performance.

We perform extensive experiments on the recently released HarperValleyBank corpus \cite{wu2020harpervalleybank} which is a task oriented dialog dataset. We focus on the multi-label classification task of dialog action recognition. When gold transcripts are available for fine-tuning an ASR, our model performs competitively with a state-of-the-art cascaded baseline while using 48\% fewer parameters. In the absence of gold transcripts, our model significantly outperforms this baseline.

\section{Hierarchical Conversation Model}
\label{sec:conv_model}
We treat the SLU problem as a context labeling task, where instead of labeling the current utterance, the model labels the entire context (dialog history + current utterance). Let the speech based context, $C^s = (x_1^s, x_2^s, ..., x_N^s)$ where $x_N^s$ is the current utterance and the rest is dialog history. Text based context, $C^t = (x_1^t, x_2^t, ..., x_N^t)$ which are gold transcripts of $C^s$. The model overview shown in Figure \ref{fig:model_over} is described below. \\

\noindent \textbf{Speech encoder (Student)}: The speech signal, $x_i^s$ from the $i^{th}$ utterance in the context is encoded using a neural network.
\begin{equation*}
    \begin{split}
        \textbf{u}_i^s &= f^s(x_i^s; \boldsymbol\theta^s)
    \end{split}
\end{equation*}
$f^s(.;\boldsymbol\theta^s)$ is the speech encoder parameterized by $\boldsymbol\theta^s$. The encoded speech context is denoted as, $\textbf{U}^s = \{\textbf{u}_i^s\}_{i=1}^{N}$. 

We use the transcription network of a pretrained RNN-T based ASR model \cite{saon2021advancing} as the speech encoder which encodes the speech signal directly to give a vector representation. This model is first pre-trained for ASR using 40-dimensional global mean and variance normalized log-mel filterbank features, extracted every 10 ms. These features are further augmented with $\Delta$ and $\Delta\Delta$ coefficients, every two consecutive frames are stacked, and every second frame is skipped, resulting in 240-dimensional vectors every 20 ms. Thus, $x_i^s$ is a sequence of 240-dimensional log-mel features. The encoder, $f^s(.;\boldsymbol\theta^s)$ is a 6-layer bidirectional LSTM with 640 cells per layer per direction. We concatenate vectors from the forward and backward direction of the LSTM's last layer for the last time step to get a 1280-dimensional vector. This vector goes through a linear layer which shrinks its dimensionality to 768. Thus, $\textbf{u}_i^s$ is a 768-dimensional representation of a speech utterance.\\

\noindent \textbf{Text encoder (Teacher)}: The $i^{th}$ text utterance, $x_i^t$ is a sequence of word tokens encoded using WordPiece embeddings. $x_i^t$ is encoded as,
\begin{equation*}
    \begin{split}
        \textbf{u}_i^t &= f^t(x_i^t; \boldsymbol\theta^t)
    \end{split}
\end{equation*}
$f^t(.;\boldsymbol\theta^t)$ is a text encoder parameterized by $\boldsymbol\theta^t$. The encoded text context is denoted as, $\textbf{U}^t = \{\textbf{u}_i^t\}_{i=1}^{N}$.

We use a pretrained BERT model\footnote{\texttt{https://huggingface.co/bert-base-uncased}} \cite{devlin2018bert} as the text encoder, $f^t(.;\boldsymbol\theta^t)$. The 768-dimensional [CLS] token output from this model is treated as the text representation, $\textbf{u}_i^t$. The text encoder is only used during training to transfer semantic knowledge to the speech encoder and is discarded at test time.\\

\noindent \textbf{Conversation encoder (Modality agnostic)}: Once we get representations for the utterances in a context for the speech and text modalities, we use a higher level sequence encoder, $g(.; \boldsymbol\phi)$ to encode the context into a single representation.
\begin{equation*}
    \begin{split}
        \textbf{c}^s &= g(\textbf{U}^s; \boldsymbol\phi) \\
        \textbf{c}^t &= g(\textbf{U}^t; \boldsymbol\phi)
    \end{split}
\end{equation*}
Here, $\textbf{c}^s$ and $\textbf{c}^t$ are representations for the speech and text context respectively. Notice that the sequence encoder is shared between the speech and text modalities. This helps to transfer knowledge from text to speech during training \cite{rongali2020exploring}.

We model $g(.; \boldsymbol\phi)$ as a 6-layer 1-head transformer (with a similar architecture as Vaswani et al. \cite{vaswani2017attention}). We add absolute positional embeddings to the sequence and pass it to the transformer. The output of the last step in the sequence output of the transformer is used as $\textbf{c}^s$ or $\textbf{c}^t$. We also explore an LSTM architecture for $g(.; \boldsymbol\phi)$ and show additional results.

\section{Training procedure}
\label{sec:train_pro}
The representations $\textbf{c}^s$ and $\textbf{c}^t$ from $g(.; \boldsymbol\phi)$ are passed into a classification layer to get predictions. The binary cross-entropy loss is then computed for the speech (denoted as $L_{BCE}(\boldsymbol\theta^s, \boldsymbol\phi)$) and the text modalities (denoted as $L_{BCE}(\boldsymbol\theta^t, \boldsymbol\phi)$). The proposed model can be used in three ways:
\begin{enumerate}
    \item \textbf{HIER-ST}: Co-trained with speech and text using $L_{BCE}(\boldsymbol\theta^s, \boldsymbol\phi) + L_{BCE}(\boldsymbol\theta^t, \boldsymbol\phi)$.
    \item \textbf{HIER-S}: Trained with speech only using $L_{BCE}(\boldsymbol\theta^s, \boldsymbol\phi)$.
    \item \textbf{HIER-T}: Trained with text only using $L_{BCE}(\boldsymbol\theta^t, \boldsymbol\phi)$.
\end{enumerate}

\subsection{Cross-modal loss functions}
\label{subsec:cmlf}
In general, BERT embeddings are well suited for NLU tasks as they are pretrained on large text corpora. The semantic knowledge from BERT can be transferred to speech embeddings through cross-modal loss functions.
We investigate two losses for this purpose, Euclidean loss and Contrastive loss. \\ 

\noindent \textbf{Euclidean loss} ($L_{EUC}$): This is computed as the $L2$ distance between speech and text representations, \cite{huang2020leveraging}. Formally,
\begin{equation*}
    \begin{split}
        L_{EUC}(\boldsymbol\theta^s) = \frac{1}{|B|}\sum_{i=1}^{|B|}\left\Vert \textbf{u}_N^s[i] - \textbf{u}_N^t[i] \right\Vert_2
    \end{split}
\end{equation*}
where $|B|$ is batch size; $\textbf{u}_N^s[i]$ and $\textbf{u}_N^t[i]$ are respective outputs from $f^s(.;\boldsymbol\theta^s)$ and $f^t(.;\boldsymbol\theta^t)$ of the $i^{th}$ utterance in the batch.\\

\noindent \textbf{Contrastive loss} ($L_{CON}$): We define the similarity between the $i^{th}$ speech and the $j^{th}$ text utterance in a training batch as:
\begin{equation*}
    \begin{split}
        s_{ij} = cos(\textbf{u}_N^s[i], \textbf{u}_N^t[j])/\tau
    \end{split}
\end{equation*}
where $cos(.,.)$ is the cosine similarity and $\tau$ is a temperature hyperparameter. Then, we define the contrastive loss as:
\begin{equation*}
    \begin{split}
    L_{CON}(\boldsymbol\theta^s) = \\
     -\frac{1}{2|B|}\sum_{i=1}^{|B|}(&\log\frac{\exp(s_{ii})}{\sum_{j=1}^{|B|} \exp(s_{ij})} + \log\frac{\exp(s_{ii})}{\sum_{j=1}^{|B|}\exp(s_{ji})})
    \end{split}
\end{equation*}
The first term in the above sum is the log-likelihood of a text utterance given the corresponding speech and the second term, vice-versa. By minimizing the above loss, the similarity between same utterances in different modalities is maximized and that between different utterances is minimized.

The gradients from these losses only update the speech branch so that the BERT representations are not affected.

All models were trained in PyTorch on a A100 GPU using the Adam optimizer at a learning rate of $2e-5$. We use 10\% dropout and early stopping with a patience of 10 epochs. The maximum context length is set to 10 utterances. For $L_{CON}$, we set $\tau = 0.07$ after tuning it on the development set.

\subsection{DropFrame}
\label{sec:df}
Compared to a sequence of text tokens, we noticed that the corresponding sequence of speech frames can be five times longer. When dialog history is used in speech form, this sequence can become very long which results in an increased training time for E2E models.


To solve this problem, we propose a simple yet effective novel technique that we call DropFrame. If the length of the sequence, $x_i^s$ is longer than a hyperparameter $l$, we randomly drop out some of the frames in $x_i^s$ such that its length becomes $l$. In our experiments, we found that $l=256$ works best. At test time, we switch off DropFrame. 
This technique not only decreases the training time but also improves performance.

\section{Experiments}
\label{sec:exp}

\subsection{Dataset}
\label{subsec:data}
We conduct experiments on the HarperValleyBank (HVB) corpus \cite{wu2020harpervalleybank} which is a public dataset of simulated spoken conversations between bank employees and customers. It consists of 1446 dyadic human-human conversations with audio and human annotated transcripts. There are 59 unique speakers involved and the data is annotated for speaker identity, caller intent, dialog actions and emotional valence.

In this work, we focus on the dialog action recognition task which is a multilabel classification problem. There are 16 dialog actions and each utterance can be labelled with one or more of these. We split the conversations into train, development and test sets. The train set contains 1174 conversations (10 hours of audio) and the test set has 199 conversations (1.8 hours hours of audio), the rest make up the development set\footnote{To get a conversation level data split, it was necessary to accept speaker overlap between the train, development, and test sets.}.

\subsection{Effect of DropFrame}
\label{subsec:eff_df}
\begin{table}
    \centering
    \resizebox{0.6\columnwidth}{!}{
    \begin{tabular}{|c|c|c|}
    \hline
    \textbf{\# Frames ($l$)} & \textbf{Macro-F1} & \textbf{Train time} \\
    \hline
    64 & 56.5 & 4 \\
    \textbf{256} & 61.7 & 8 \\
    1024 & 60.2 & 25 \\
    All & 59.9 & 27 \\
    \hline
    \end{tabular}}
    \caption{Effect of DropFrame. Train time is in minutes/epoch.}
    \label{tab:table_3}
\end{table}

Table \ref{tab:table_3} shows the effect of our proposed DropFrame approach (Section \ref{sec:df}). We fine-tuned the frame length $l$ with several values on the development set, but show results for only four of these values on the test set. As we can see, the training time is reduced significantly when frames are dropped. Also note that when we increase the value of $l$ beyond 256, the performance starts to drop. We hypothesize that dropping out frames works as an effective regularizer which helps the model generalize better by aggregating over different sequences of frames for the same utterance \cite{park2019specaugment, jain2021spliceout}. A more thorough analysis of this is left for future work.

\subsection{SLU results}
\label{subsec:results}
\begin{table}
    \centering
    \resizebox{\columnwidth}{!}{
    \begin{tabular}{|cl|c|c|}
    \hline
    & \textbf{Model} & \textbf{Macro-F1} & \textbf{\# Params}\\
    \hline
    \textbf{(1T)} & BERT (on utterance) & 56.1 & 168M \\
    \textbf{(2T)} & BERT (on context) & \textbf{63.5} & 168M \\
    \textbf{(3T)} & HIER-T & 63.3 & 200M \\
    \hline
    \textbf{(1C)} & ASR $\longrightarrow$ BERT (on context) & \textbf{62.2} & 168M \\
    \textbf{(2C)} & ASR $\longrightarrow$ HIER-T & 61.3 & 200M \\
    \hline
    \textbf{(1E)} & LSTM (on utterance) & 54.0 & 54M \\
    \textbf{(2E)} & HIER-S & 58.3 & 88M \\
    \textbf{(3E)} & HIER-ST & 59.0 & 88M \\
    \textbf{(4E)} & HIER-ST + $L_{EUC}$ & 60.3 & 88M \\
    \textbf{(5E)} & HIER-ST + $L_{CON}$ & \textbf{61.7} & 88M \\
    \textbf{(6E)} & HIER-ST + $L_{EUC}$ + $L_{CON}$ & 60.9 & 88M \\
    \textbf{(7E)} & HIER-ST + $L_{CON}$ ($g(.;\boldsymbol\phi)$ = LSTM) & 61.3 & \textbf{62M} \\
    \hline
    \end{tabular}}
    \caption{Performances when gold transcripts are available with model size (in million parameters). \textbf{(T)} denotes text-based models; \textbf{(C)} denotes ASR-NLU cascaded models; \textbf{(E)} denotes E2E models. In \textbf{(C)}, the ASR model is adapted on in-domain data. In \textbf{(E)}, the speech encoder is the transcription network from an adapted RNN-T ASR. Note that we did not observe significant variance in results between runs.}
    \label{tab:table_1}
\end{table}

\begin{table}
    \centering
    \resizebox{0.8\columnwidth}{!}{
    \begin{tabular}{|cl|c|}
    \hline
    & \textbf{Model} & \textbf{Macro-F1}\\
    \hline
    \textbf{(3C)} & ASR $\longrightarrow$ BERT (on context) & 50.3\\
    \hline
    \textbf{(8E)} & HIER-S & 57.7\\
    \textbf{(9E)} & HIER-ST + $L_{CON}$ (w/ ASR text) & \textbf{60.3}\\
    \hline
    \textbf{(10E)} & HIER-ST + $L_{CON}$ (w/ Gold text) & \textbf{61.7}\\
    \hline
    \end{tabular}}
    \caption{Performances when gold transcripts are unavailable (except \textbf{(10E)}). In \textbf{(C)}, we use an off-the-shelf ASR. In \textbf{(E)}, the speech encoder is from an off-the-shelf RNN-T ASR.}
    \label{tab:table_2}
\end{table}

Table \ref{tab:table_1} shows experiments assuming gold transcripts are available during training. \textbf{(1T-3T)} shows the oracle performance on clean text data. In \textbf{(1T)}, only the current utterance is fed to BERT whereas in \textbf{(2T)}, the entire context (dialog history + current utterance) is given to BERT. \textbf{(3T)} is our hierarchical model with only the text branch. Clearly, using dialog history on text makes a big difference in performance. In the cascaded ASR-NLU settings, \textbf{(1C-2C)}, we see a drop in performance compared to \textbf{(2T-3T)}, because of ASR errors.

\textbf{(1E-7E)} represent E2E models using a pretrained ASR based speech encoder. This encoder corresponds to a 6-layer LSTM transcription network of an RNN-T model adapted on the HVB dataset. \textbf{(1E)} uses this LSTM on the current utterance, whereas all the others use dialog history in speech form through our hierarchical setup. Note that \textbf{(2E)} already gives a 4.3\% improvement over \textbf{(1E)} which shows that our hierarchical model effectively uses history in an E2E manner. \textbf{(3E-7E)} use the available text transcriptions to train the full model, although the text branch is discarded at test time. We see that co-training with text helps improve performance. By adding the different cross-modal losses defined in Section \ref{subsec:cmlf}, we see further performance improvements. Note that our best performing E2E model, \textbf{(5E)} uses the proposed contrastive loss. This model shows comparable performance to \textbf{(1C)} which has a distinct advantage that BERT can attend to the entire text conversation in a more fine-grained manner at the word level. This is not feasible in a speech based conversation which can be intractably longer compared to text. Hence, we have proposed a tractable hierarchical setup. Notably, \textbf{(5E)} gives a better performance compared to a similar hierarchical cascaded setup, \textbf{(2C)} which shows its robustness to ASR errors.

Note that all of the proposed end-to-end models are significantly smaller in size compared to the cascaded models. To mimic a more deployable scenario, we replaced the transformer-based conversation encoder, $g(.;\boldsymbol\phi)$ with a 1-layer bidirectional LSTM and add the forward and backward representations of the last time step to get the context representations. The result is shown in row \textbf{(7E)}. We see that the F1 is only 0.9\% less than the cascaded baseline \textbf{(1C)}, but the size of the former is 64\% smaller and hence easily deployable. This also shows that the conversation encoder is agnostic to the underlying neural architecture.

For the next set of experiments in Table \ref{tab:table_2}, we assume a constrained SLU setting. Here, gold transcripts are not available to adapt the ASR model. Instead, only the SLU labels are available. This practical setting is common as it reduces the transcription cost when creating an SLU dataset. The cascaded baseline in this case \textbf{(3C)} has a much worse performance compared to \textbf{(1C)} in Table \ref{tab:table_1}. As the ASR used in \textbf{(3C)} has not been adapted to the in-domain data, its output can be noisy (WER of 11.3\% compared to the adapted case with a WER of 1.7\%). We also observe that in a lot of cases the unadapted RNN-T based ASR model does not output a transcript because of acoustic mismatch. This can be detrimental for the cascaded SLU system. In the absence of human transcriptions, we trained our best performing HIER-ST model from Table \ref{tab:table_1} (with $L_{CON}$) with the ASR hypotheses from the off-the-shelf ASR. The result is shown in row \textbf{(9E)} and we observe a significant improvement of 10\% absolute F1 over \textbf{(3C)}. This is because our model is fully E2E and robust to ASR errors.

We also conduct an additional experiment where although human transcriptions are available for the spoken data, they are not used for adapting the ASR; rather, they are used to co-train the HIER-ST model only \textbf{(10E)}. In this setup, we see a similar performance as \textbf{(5E)} in Table \ref{tab:table_1}. This shows that our model's speech encoder does not need to be adapted to every new domain using the ASR objective and can be used off-the-shelf. This can save a lot of time in new domain deployments. 

\section{Conclusion}
We believe that this work is the first step towards building a fully E2E SLU system using dialog history in speech form. We propose a hierarchical conversation model that can encode speech utterances at the lower level and the entire conversation at the higher level. We use text transcriptions to co-train a similar text-based system which acts as a teacher for our E2E system. We also incorporate cross-modal loss functions which further enhance the performance of the E2E system. Future work should explore a computationally efficient encoding of speech-based conversations using a single BERT-like model. 

\vfill\pagebreak

\bibliographystyle{IEEEbib}
\bibliography{strings,refs}

\end{document}